%% file: icdp2009.tex
\documentclass[10pt]{article}

\input{preamble.tex}

\usepackage{lmodern}

\usepackage{graphicx}
\usepackage{times}
\usepackage{algorithm}
\usepackage{algorithmic}

\begin{document}
\noindent

\bibliographystyle{ieeetr}

\title{Anomaly Detection Based on Multiple-Hypothesis Autoencoder}

\authorname{JoonSung Lee \thanks{Correspondence author: js43.lee@sk.com}, YeongHyeon Park}
\authoraddr{SK Planet Co., Ltd.}

\maketitle


\abstract
Recently Autoencoder(AE) based models are widely used in the field of anomaly detection.A model trained with normal data generates a larger restoration error for abnormal data. Whether or not abnormal data is determined by observing the restoration error. It takes a lot of cost and time to obtain abnormal data in the industrial field. There- fore the model trains only normal data and detects abnormal data in the inference phase. However, the restoration area for the input data of AE is limited in the latent space. To solve this problem, we propose Multiple-hypothesis Autoencoder(MH-AE) model composed of several decoders. MH-AE model increases the restoration area through contention between decoders. The proposed method shows that the anomaly detection performance is improved com- pared to the traditional AE for various input datasets.

\section{Introduction}
\label{sec:introduction}
Recently, anomaly detection has been used in various industrial fields 
such as finance, medicine, manufacturing, and security in combination with deep learning.
Anomaly detection refers to finding data that has a large difference from normal data.
However, collecing abnormal data in the industrial field is costly and time consuming.
In order to solve for these limitations, only normal data are trained in the train pahse, 
and abnormal data are detected in the test phase \cite{7}\cite{9}.

However, the restoration range of AE for the input data of is limited in the latent space.  

In this paper, we propose Multiple-hypothesis Autoencoder (MH-AE) model to compensate for these limitations.
The proposed model maximizes anomaly detection performance through contention of several decoders.

\section{Related work}
\label{sec:related_work}
It takes a lot of cost and time to obtain abnormal data in the industrial field.
To ease for these imitations, the model is trained on normal data in the learning stage.
And a threshold is determined based on the distribution of the loss function
There was a research to detect abnormal data through a larger residual error in the test stage \cite{7}\cite{9}.

The restoration area for the input data of AE is limited in the latent space.  
The proposed MH-AE shows more improved reconstruction performance through several decoders.

Recently an anomaly detection model based on a multi-hypothesis neural network was published  \cite{3}\cite{4}.
Hypothesis Pruning Generative Adversarial Network(HP-GAN) that is based on Winner-Take-Al theory consists of an adversarial neural network 
trained through matching of multiple hypotheses and latent vectors.
HP-GAN has a discriminator, many parameters are used when implementing the model, 
but MH-AE shows similar performance with fewer parameters.

\section{Proposed approach}
\label{sec:proposed_approach}
In this section, we present MH-AE architecture and anomaly detection method.
The model consists of one encoder and several decoders.
The model output is that the predicted value of the decoder with the smallest loss function.

\subsection{Multiple-hypothesis autoencoder model}
\label{subsec:single_module}
\begin{figure}[ht]
    \begin{center}
		\includegraphics[width=0.9\linewidth]{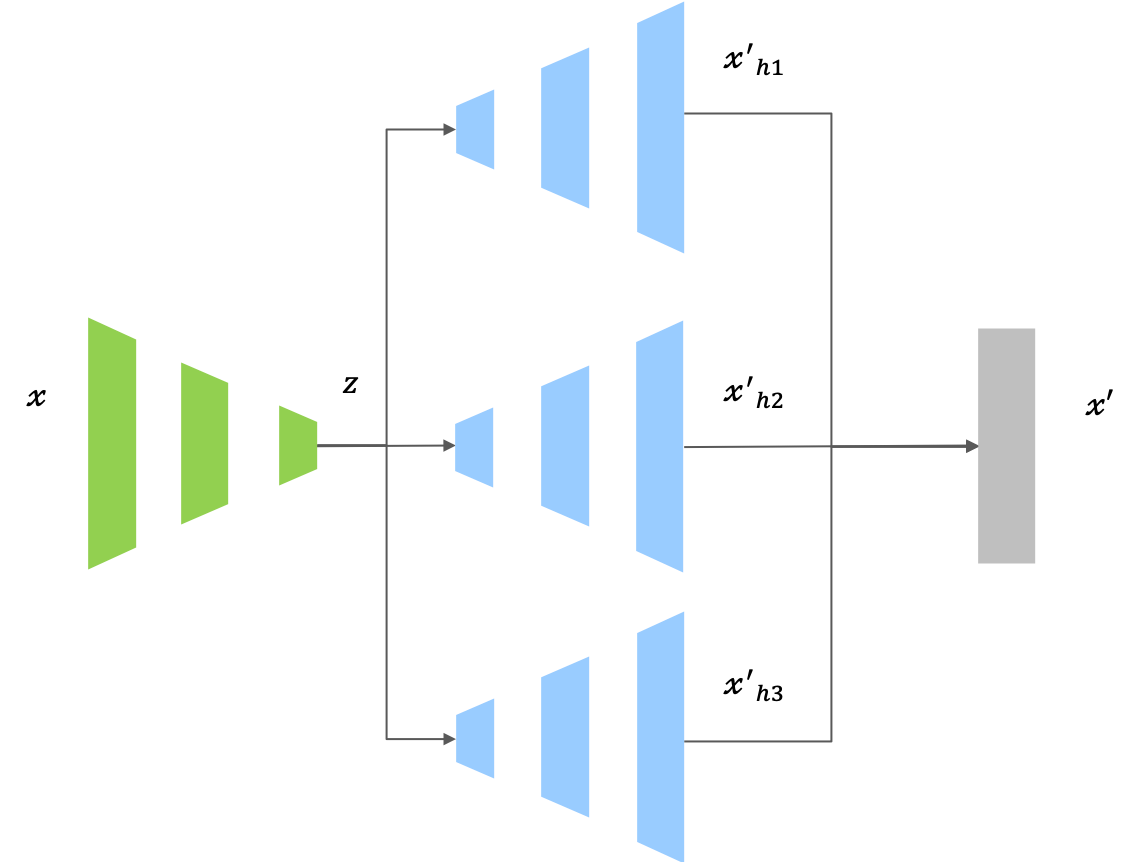}
	\end{center}
	\caption{Multiple-hypothesis autoencoder architecture.}
	\label{fig:fig1}
\end{figure}

Figure~\ref{fig:fig1} shows the architecture of MH-AE model proposed in this paper.
The proposed model consists of one encoder, three decoders, 
and a final output determination unit.
The same latent vector is applied to the input of each decoder. 
Table~\ref{tab:table1} shows the meaning of the symbols used in Figure~\ref{fig:fig1}.

To detect anomalies, first, a loss function is obtained 
using the input data of the encoder and the outputs restored by each decoder.
Among them, the output of the decoder, where the smallest loss function is observed,
is selected and sent as the final output signal. 
The loss function uses MSE.
Each time a new prediction is performed, it selects the output of the decoder 
with the best performance for each input, Anomaly detection performance is maximized

\begin{table}[ht]
    \centering
    \caption{The symbol description of MH-AE.}
    \begin{tabular}{l|l}
        \hline
            \textbf{Symbol} & \textbf{Description} \\       
        \hline
            $x$ & Input data of the model \\
            $z$ & Latent vector \\
            $x`h1$ & Output of the decoder1 \\
            $x`h2$ & Output of the decoder2 \\
            $x`h3$ & Output of the decoder3 \\            
            $x'$ & Output of the model \\
        \hline
    \end{tabular}
    \label{tab:table1}
\end{table}

\subsection{Anomaly detection method}
\label{subsec:swe}
To verify the anomaly detection performance, the unimodal MNIST dataset was used.
Training was conducted under the assumption of one class as normal data.
The anomaly detection performance was verified through the MSE 
generated by restoring all classes in the inference process.

In the training phase, the loss functions of each decoder were all equally applied to MSE, 
but the optimal parameters of each decoder were updated independently.
In the test phase, the MSE between the decoder output and the encoder input is calculated.
And the decoder with the smallest MSE is selected as the final output.
Each time prediction is performed, each decoder competes with each other,
and the output of the decoder with the least reconstruction error is adopted.

\begin{equation} 
\theta = \mu \pm 1.5  \sigma 
\end{equation}

$\theta$ in Equation 1 is determined during the training process 
and is a threshold that classifying abnormal and abnormal data.
For the MSE of $x$ and $x^{'}$, $u$  is the mean and $\sigma$  is the standard deviation.
In the test step, if the MSE of $x$ and $x^{'}$ was less than $\theta$ it was classified as normal,
otherwise it was classified as abnormal.
$\theta$ can be adjusted according to the test environment.

\section{Experiments}
\label{sec:experiments}
In this section, an experiment to confirm MH-AE based anomaly detection effect proposed in this paper is dealt with.

\subsection{Experiment result}
\label{subsec:environment}
Anomaly detection experiments were conducted 
on the MNIST data set using AE and MH-AE.
We trained the dataset corresponding to class ‘1’ with normal data.
and observed the restoration error for all classes '0~9'.
The experiment was repeated 30 times, referring to the Monte Carlo method.[8]

\begin{figure}[ht]
    \begin{center}
		\includegraphics[width=0.9\linewidth]{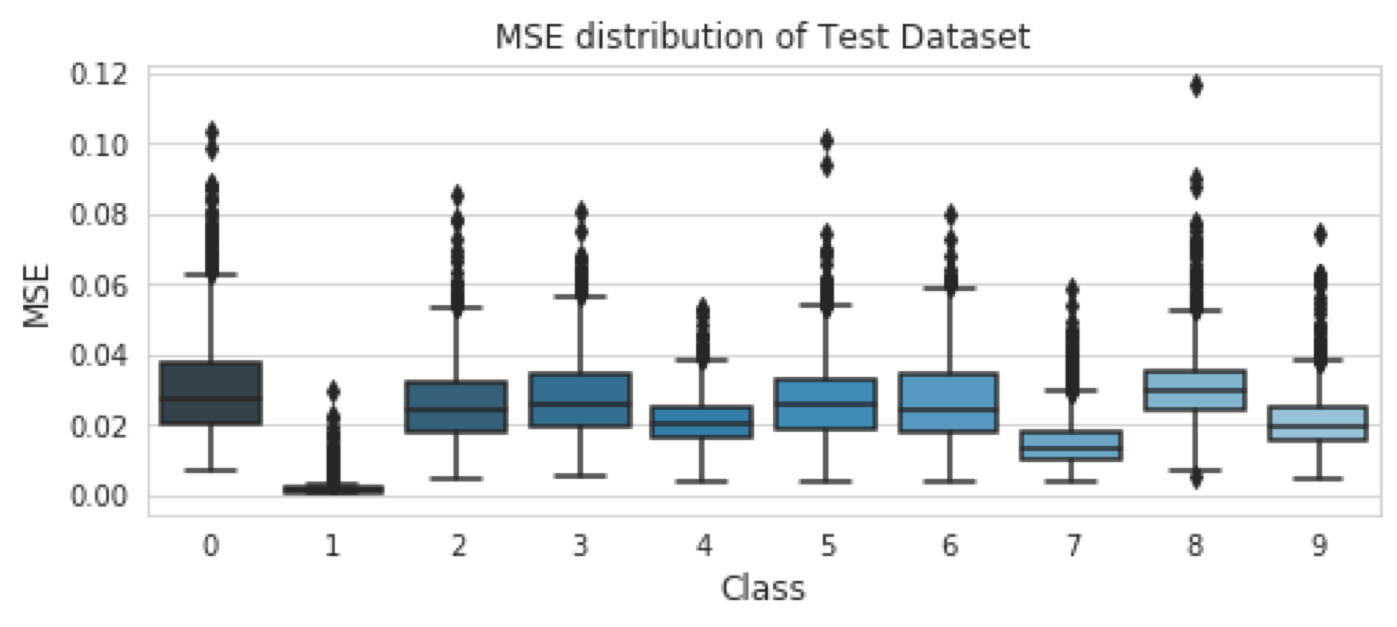}
	\end{center}
	\caption{MSE distribution for the test dataset of the model trained with ‘1’ as normal.}
	\label{fig:fig2}
\end{figure}

Figure~\ref{fig:fig2} shows the distribution of MSE for the predicted class '0~9' in the test stage.
The MSE distribution of class '1' was located in a relatively low area.
In class “7,” which is similar to “1,” the MSE distribution was 
most similar to the '1' distribution compared to other abnormal classes.

\begin{figure}[ht]
    \begin{center}
		\includegraphics[width=0.9\linewidth]{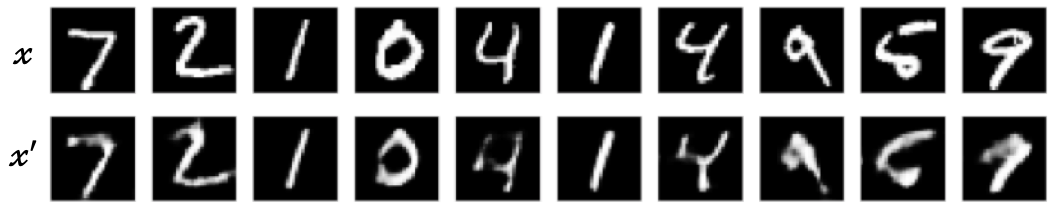}
	\end{center}
	\caption{Reconstruction image of the test dataset.}
	\label{fig:fig3}
\end{figure}

Figure~\ref{fig:fig3} shows the input and reconstructed images of the test data set.  
For the normal data '1' class, it was clearly restored,
bug for the abnormal data, the restoration area was partially limited.

\subsection{Comparison of performance}
\label{subsec:classification}
Table~\ref{tab:table2} is the final result of performing anomaly detection through the MNIST dataset.Referring to the experimental results, the anomaly detection performance of the proposed model is improved compared to AE model.

\begin{table}[ht]
    \centering
    \caption{Comparison of anomaly detection performance.}
    \begin{tabular}{l|rr}
        \hline
            \textbf{Model} & \textbf{AUROC} & \textbf{MSE} \\       
        \hline
            AE      & 0.99779  & \textbf{0.00220} \\
            MH-AE   & 0.99803  & \textbf{0.00215} \\
        \hline
    \end{tabular}
    \label{tab:table2}
\end{table}

\section{Conclusion}
\label{sec:conclusion}
In this paper, we proposed MH-AE model that shows superior anomaly detection performance
compared to traditional AE for various input dataset.
Experimentally verified that the proposed model improves the anomaly detection performance for the MNIST dataset.

\section*{Acknowledgements}
A preliminary version of the paper was presented at the 33rd Workshop on Image Processing and Image Understanding.


\end{document}

%% file: preamble.tex
\usepackage[a4paper,textwidth=18cm,textheight=24cm,top=2.85cm, bottom=2.85cm, left=1.5cm, right=1.5cm]{geometry}

\usepackage{icdp2009}

\makeatletter
\long\def\@makecaption#1#2{%
\vskip\abovecaptionskip
\sbox\@tempboxa{#1. #2}%
\ifdim \wd\@tempboxa >\hsize
#1. #2\par
\else
\global \@minipagefalse
\hb@xt@\hsize{\box\@tempboxa\hfil}%
\fi
\vskip\belowcaptionskip}
\makeatother